\documentclass[11pt,a4paper]{article}
\usepackage[hyperref]{eacl2021}
\usepackage{times}
\usepackage{latexsym}

\usepackage{times}
\usepackage{url}
\usepackage{latexsym}
\usepackage{amsmath}
\usepackage{amssymb}
\usepackage{latexsym}
\usepackage{graphicx}
\usepackage{multirow}
\usepackage{breqn}

\usepackage{tabularx}

\usepackage{graphicx}
\usepackage{amsmath,graphicx}
\usepackage{times}
\usepackage{latexsym}
\usepackage{times}
\usepackage{latexsym}
\usepackage{caption}
\usepackage{subcaption}
\usepackage{comment}
\usepackage{times}
\usepackage{soul}
\usepackage{booktabs}
\usepackage{url}
\usepackage[utf8]{inputenc}
\usepackage{graphicx}
\usepackage{amsmath}
\usepackage{amsmath,graphicx}
\usepackage{times}
\usepackage{latexsym}
\usepackage{times}
\usepackage{latexsym}
\usepackage{comment}
\usepackage{times}
\usepackage{soul}
\usepackage{booktabs}
\usepackage{url}
\usepackage[utf8]{inputenc}
\usepackage{graphicx}
\usepackage{amsmath}
\usepackage{amssymb}
\usepackage{array}
\usepackage{booktabs}
\usepackage{multirow}
\usepackage{multicol}
\usepackage{caption}
\usepackage{xspace}
\usepackage{amssymb}
\usepackage{booktabs}

\usepackage{multirow}
\usepackage{multicol}
\usepackage{caption}
\usepackage{xspace}
\usepackage{amsmath}

\usepackage{tabularx}

\usepackage{latexsym}
\usepackage{multirow}
\usepackage{algorithm} 
\usepackage{algpseudocode} 
\usepackage{multicol}
\usepackage{comment}
\usepackage{hyperref}
\usepackage{amsmath}

\usepackage{tabularx}

\usepackage{graphicx}
\usepackage{subcaption}
\usepackage{url}
\DeclareSymbolFont{rsfs}{U}{rsfs}{m}{n}
\DeclareSymbolFontAlphabet{\mathscrsfs}{rsfs}
\usepackage{mathtools}

\DeclarePairedDelimiter\abs{\lvert}{\rvert}%
\DeclarePairedDelimiter\norm{\lVert}{\rVert}%

\makeatletter
\let\oldabs\abs
\def\abs{\@ifstar{\oldabs}{\oldabs*}}
\let\oldnorm\norm
\def\norm{\@ifstar{\oldnorm}{\oldnorm*}}
\makeatother
\usepackage{microtype}

\aclfinalcopy 
\def\aclpaperid{1378} 

\title{WER--BERT: Automatic WER Estimation with BERT in a Balanced Ordinal Classification Paradigm}

\author{Akshay Krishna Sheshadri* \\
\\\And
Anvesh Rao Vijjini* \\ \\
 Samsung R\&D Institute India - Bangalore \\
 \texttt{\{a.sheshadri,a.vijjini,sukhdeep.sk\}@samsung.com} \\\And
 Sukhdeep Kharbanda \\
 \\}

\date{}

\begin{document}
\maketitle
\begin{abstract}
Automatic Speech Recognition (ASR) systems are evaluated using Word Error Rate (WER), which is calculated by comparing the number of errors between the ground truth and the transcription of the ASR system. This calculation, however, requires manual transcription of the speech signal to obtain the ground truth. Since transcribing audio signals is a costly process, Automatic WER Evaluation (e-WER) methods have been developed to automatically predict the WER of a speech system by only relying on the transcription and the speech signal features. While WER is a continuous variable, previous works have shown that positing e-WER as a classification problem is more effective than regression. However, while converting to a classification setting, these approaches suffer from heavy class imbalance. In this paper, we propose a new balanced paradigm for e-WER in a classification setting. Within this paradigm, we also propose WER-BERT, a BERT based architecture with speech features for e-WER. Furthermore, we introduce a distance loss function to tackle the ordinal nature of e-WER classification. The proposed approach and paradigm are evaluated on the Librispeech dataset and a commercial (black box) ASR system, Google Cloud’s Speech-to-Text API. The results and experiments demonstrate that WER-BERT establishes a new state-of-the-art in automatic WER estimation.
\end{abstract}

\section{Introduction}
\label{intro}
\let\thefootnote\relax\footnotetext{*The authors contributed equally to the work.}

ASR systems are ubiquitous now. They are available across applications such as Voice Assistants, Assisted Living or Hands free device usage. However, with the widespread usage of ASR systems, there comes a heavy need for ASR Evaluation as well - to select, compare or improve alternate ASR systems. WER is widely considered as the standard metric for ASR evaluation. A higher WER means a higher percentage of errors between the ground truth and the transcription from the system. WER is calculated by aligning the two text segments using string alignment in a dynamic programming setting. The formula is as follows:
\begin{equation}
\label{eq:1}
  WER = \frac{ERR}{N}
\end{equation}

\begin{figure*}
\centering
 \includegraphics[width=12cm]{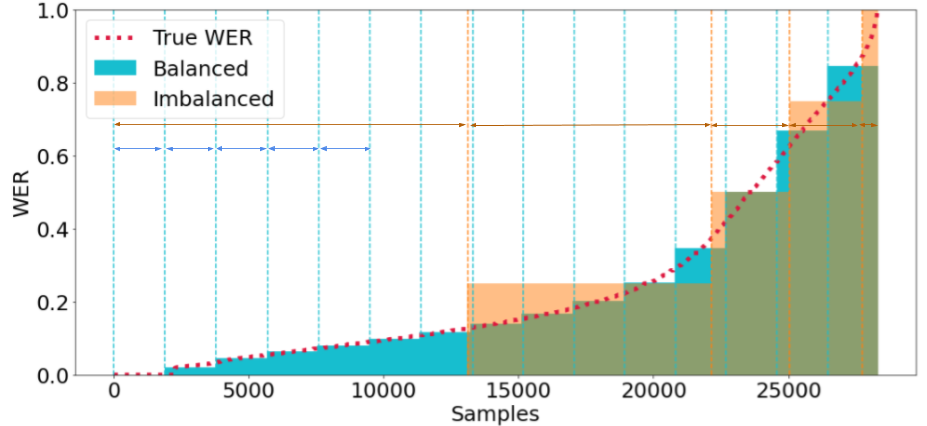}
 \caption{Distribution of WER classes of the 100 hr dataset, using proposed Balanced approach and using the \protect\newcite{elloumi2018asr} imbalanced approach. The space between two consecutive vertical lines indicates the size of a respective class. The blue lines are evenly spaced, whereas the orange lines are spaced irregularly, indicating an imbalanced distribution.}
 \label{fig:bal}
\end{figure*}
where $ERR$ is the sum of errors (Insertions, Deletions or Substitutions) between the transcription and the ground truth. $N$ is number of words in ground truth. As evident from this equation, the presence of ground truth is necessary for the calculation of errors, and hence, for WER.
However, manual transcription of speech at word level is a expensive and dilatory process. Hence, the need for an automatic ASR evaluation is important but few attempts have addressed this. Furthermore, for effectively training, evaluating and judging the performance of an ASR system, WER calculation needs to be done on adequate hours of data. As this test set increases, a more accurate estimation of WER is possible. Since automatic WER evaluation does not have the bottleneck of manual transcription, it can be calculated over large test sets leading to more accurate estimates. 
Because of the immense popularity of attention based architectures for text classification \cite{madasu2019sequential,choudhary2020self,madasu2020position,rao2020sentiwordnet,madasu2020sequential}, we propose the transformer \cite{vaswani2017attention} encoder architecture --- Bidirectional Encoder Representations from Transformers (BERT) \cite{devlin2018bert} for e-WER. BERT is pretrained on huge amounts of open domain language and is extensively used for its effectiveness in natural language understanding tasks. By pretraining on such data, the model gains knowledge of general domain language structure which aids in predicting speech errors which are typically observed as a deviation from the general syntax and semantics of a sentence. While previous approaches address e-WER \cite{ali2018word,elloumi2018asr,elloumi2019investigating} in classification settings, their models suffer with gross imbalance in the WER classes. To address this issue, we present a training framework which will always consist of training on equal sized classes no matter the true WER distribution. 
Additionally, these previous e-WER classification tasks assume that there is no inherent relative ordering to the classes. However, this is not the case with WER classification since a misclassification closer to the ground truth will lead to a lower mean absolute error (MAE) in the WER prediction. Such classification tasks are called Ordinal classification\cite{frank2001simple}.

The overall contributions of our paper can be summarized follows:

(i) A new balanced paradigm for training WER as a classification problem which addresses the label imbalance issue faced by previous works.

(ii) WER-BERT - We find that language model information is helpful in predicting WER and hence propose a BERT based architecture e-WER. 

(iii) Distance Loss to address the ordinal nature of WER classification which penalizes misclassifications differently based on how far off the predicted class is when compared to the real class.

\section{Related Work}
 \label{related}

While the importance of an automatic WER prediction system is immense, there have not been many works directly addressing it. Related works such as exploring the word-level confidence in ASR prediction are abundant \cite{seigel2011combining,huang2013predicting}. There have also been works predicting the errors or error estimates as well in some form \cite{ogawa2015asr,ogawa2017error,seigel2014detecting,yoon2010predicting}. These approaches either predict some of the errors described in WER prediction or alternate metrics to rate ASR systems such as accuracy or error type classification. However, they lack calculation of the complete WER score. Transcrater \cite{jalalvand2016transcrater} was one of the first works which aim at predicting WER directly. They propose a neural network in a regression setting trained on various features such as parts of speech, language model, lexicon, and signal features. However, more recent approaches\cite{ali2018word,elloumi2018asr,elloumi2019investigating} phrase WER prediction as a classification approach. \newcite{ali2018word} propose two types of models based on the input available --- the glassbox model which uses internal features of the target ASR system such as its confidence in transcribing the audio clip; and the black box model which only uses the transcripts and other features generated from the transcript such as the word and the grapheme count. They propose a bag-of-words model along with additional transcription features such as duration for e-WER.

The black box setting is a harder task since ASR model features such as the average log likelihood and the transcription confidence can give a good indication on how many errors may have occurred during the automatic transcription. However, the black box approach is not specific to the architectural design of an ASR system and can be used with any ASR system without access to its internal metrics such as the aforementioned confidence. Thus our proposed approach is trained in a black box setting.

\newcite{elloumi2018asr,elloumi2019investigating} build a CNN based model for WER classification. We built models based on them as baselines to evaluate WER-BERT's performance. They are further explained in the Sections \ref{sec:singlevsdouble} and \ref{sec:baselines}.

ASR errors often make a transcription ungrammatical or semantically unsound. Identifying such constructs is also reflected in the dataset of Corpus of Linguistic Acceptability(CoLA) \cite{warstadt2019neural,warstadt2019cola}. CoLA is a dataset intended to gauge at the linguistic competence of models by making them judge the grammatical acceptability of a sentence. CoLA is also part of the popular GLUE benchmark datasets for Natural Language Understanding \cite{wang2018glue}. BERT \cite{devlin2018bert} is known for outperforming previous GLUE state-of-the-Art models, including the CoLA dataset.

\section{Dataset}
For our experiments, we have used the Librispeech dataset \cite{panayotov2015librispeech} which is a diverse collection of audio book data along with the ground text. It has around 1000 Hours of audio recordings with different levels of complexity. We pass these audio clips through an ASR system to get its transcripts and the WER is calculated by comparing it with the ground text. This paper reports findings in the experiments run with Google Cloud's Speech-to-Text API\footnote{https://cloud.google.com/speech-to-text.}. We chose this commercial ASR system, rather than reporting results on an internal ASR system, since it's easily accessible through the google-api and the results are reproducible. For our experiments, we have used the 10 and 100 hour datasets and made a 60:20:20 split into train, dev and test sets for each dataset. As can be seen in Table 1 of Section B of the Appendix, the characteristics of the 100 and 10 hour datasets are quite different. However, within a dataset, the train, dev and test sets have similar distributions of WER and other characteristics. Table \ref{tab:wer_examples} lists a few examples with each row having the ground text, the transcript obtained from Google Cloud's Speech-to-Text API, the True WER and the WER predicted by our proposed model.

\begin{figure}
\centering
 \includegraphics[width=7.5cm]{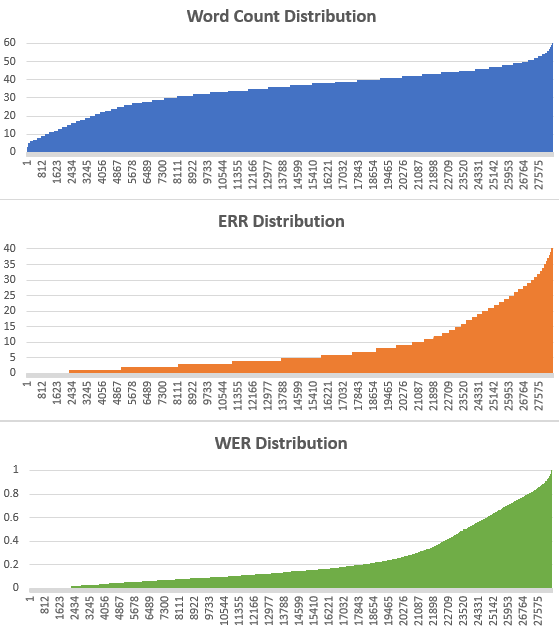}
 \caption{Distribution of WER, ERR and Word Count on our Training (Train 100hr) set.}
\label{fig:dist}
\end{figure}

\section{Single Task and Double Task for WER Estimation}
\label{sec:singlevsdouble}
As shown in Equation \ref{eq:1}, the WER of an utterance is the fraction obtained by the division of 2 integers --- Errors per sentence (ERR), which is the total number of insertions, deletions and substitutions needed to convert an ASR's transcript to the ground text, and the word count of the ground text (N). Since the WER of a sentence is a continuous variable between 0 and 1 (mostly), a common way to model this is through a regression model. \newcite{elloumi2018asr,elloumi2019investigating} instead present a way to turn this into a classification problem for e-WER. They experiment with various combinations of text and audio signal inputs and show that a classification approach outperforms its corresponding regression approach trained on the same inputs. \newcite{elloumi2018asr}'s approach estimates WER directly with a 6 class classification model (with classes corresponding to $0\%, 25\%, 50\%, 75\%, 100\%$ and $150\%$). Once the model is trained, the predictions are calculated as follows:
\begin{equation}
\label{eq:dot}
  WER_{Pred}(s) = P_{softmax}(s) \cdot WER_{fixed}
\end{equation}
where $s$ is a sample, $WER_{Pred}$ is the predicted WER for $s$, $P_{softmax}(s)$ is the softmax probability distribution 
output by the classification model,
\begin{math} WER_{fixed} = [ 0, 0.25, 0.5, 0.75, 1.0, 1.5] \end{math} is the fixed vector and `$\cdot$' is the dot product operator. We call this approach as the Single Task method for e-WER. Alternatively, \newcite{ali2018word} present another classification approach for e-WER. They argue that the calculation of WER relies on two distinct calculations --- ERR and N. Since both of them are discrete integers, they propose two independent classification problems to the estimate errors in the sentence $ERR_{est.}$ and to estimate the Word Count of the ground text $N_{est.}$. The predicted WER is then calculated as $(ERR_{est.}/N_{est.})$. We call this approach as the Double Task method for e-WER.

\begin{figure*}
\centering
 \includegraphics[width=12cm]{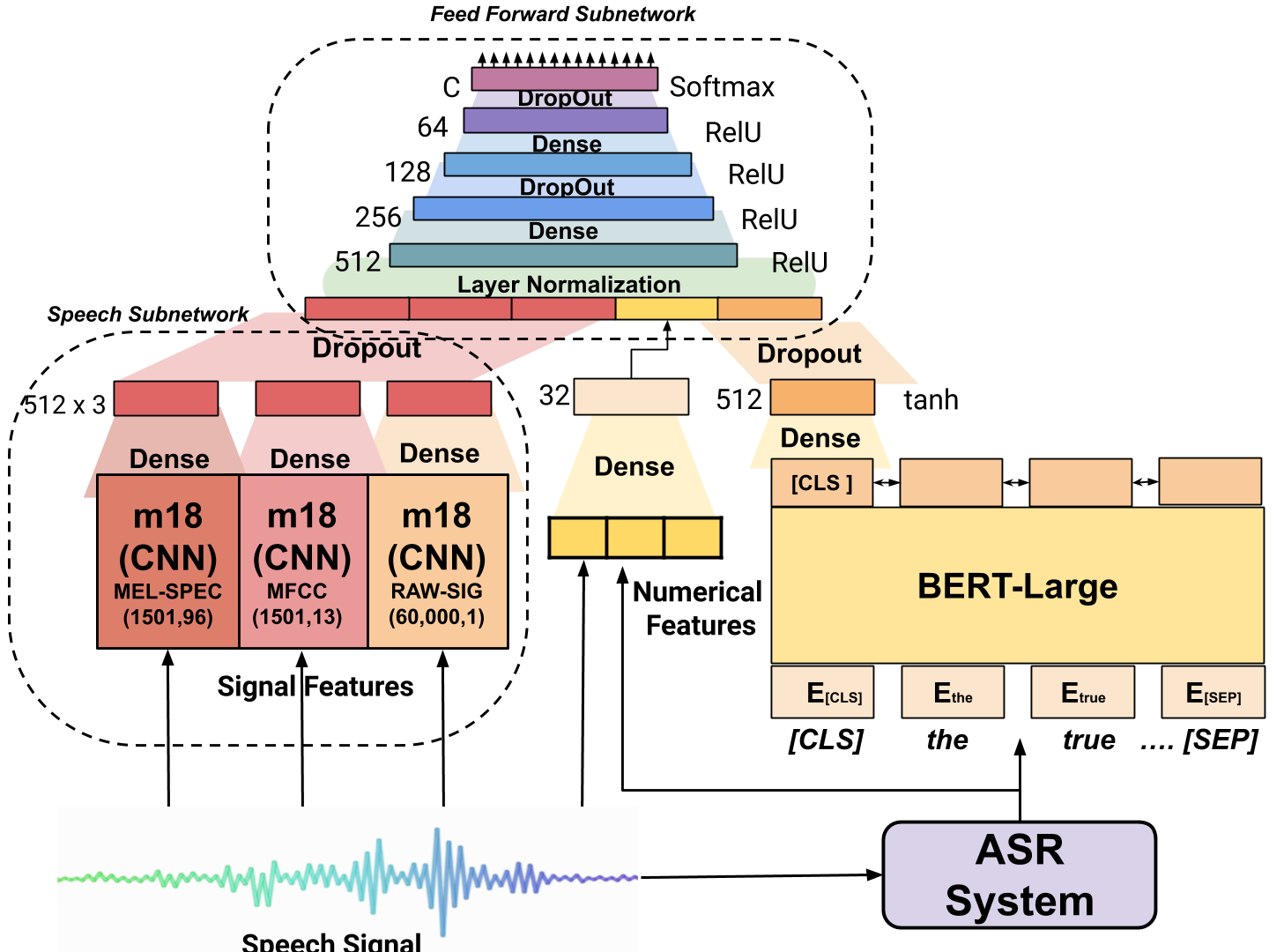}
 \caption{Proposed WER-BERT}
 \label{fig:arch}
\end{figure*}

\subsection{The Problem of Class Imbalance}
While these approaches had good results, both of them suffer from the problem of imbalanced classes. As we can seen from Figure \ref{fig:dist}, the ERR and the WER are highly imbalanced. The nature of WER is such that it is very less likely for this task to be balanced for any data and any ASR system. This imbalance leads to poor performance due to certain WER or ERR classes having very few samples in them. With \newcite{elloumi2018asr}, all the true WER's are `cast' to the nearest multiple of 0.25. It can be seen from Figure \ref{fig:dist} that the number of samples belonging to a class varies tremendously in this imbalanced setting. This leads to the model having poor performance, especially in the higher WER ranges. Moreover, different ASR systems will have their own distributions and some may be relatively well balanced in some ranges while other may be much worse. This approach is not scalable and it fails to generalize a method for creating balanced class distributions, irrespective of the ASR system.

\subsection{Proposed Balanced Division}
\label{sec:balanced}
We propose an alternate paradigm for creating WER class distributions. We extend the single task setting for e-WER to a balanced WER class distribution irrespective of the dataset and ASR system. Instead of fixing a list of WER values based on a factor such as 0.25 to represent the classes, the total number of classes desired $K$\footnote{Refer to Section A.2 of the appendix for tuning of the class hyperparameter $K$ } is decided. Let $w_{1},w_{2}, w_{3},.... w_{D}$ be the WERs of individual samples ordered in an ascending manner where $D$ is the total number of samples in the corpus. Then the number of samples in each of the $K$ classes when divided uniformly will be $n_{s} = (D/K)$. A class $C_{i}$ will be defined as the samples with WER in the range: [$w_{(i-1)*n_{s}}, w_{(i)*n_{s}}$] where $i$ $\in$ $\{1,2,...K\}$. This is shown in Figure \ref{fig:bal} where $K = 15$ classes are made with equal number of samples in each. The WER value $WER_{Fixed_i}$ associated with each class $C_{i}$ is defined as the mean WER of that class:
\newline
\begin{equation}
  WER_{Fixed_{i}} = \sum_{j =(i-1)*n_{s}}^{(i)*n_{s}} \frac{w_{j}}{n_{s}}
\end{equation}

Once $WER_{Fixed}$ is calculated, we use Equation \ref{eq:dot} to compute the predicted WER of a sample using $P_{softmax}$ calculated from a neural network model. Apart from being balanced, this approach also has the benefit of generating classes which fits the true WER curve better than the previous approach as show in Figure \ref{fig:bal}. It's important to understand that while the ERR estimation in the Double Task setting of \newcite{ali2018word} is also imbalanced, it can't be mapped into arbitrary classes based on ordering. Since WER is a continuous variable, unlike ERR, it is possible to decide the boundary for a WER class arbitrarily to create balanced sets.

\begin{table*}[h]
\small
\begin{tabularx}{\textwidth}{X X l l}
  \hline
  \hline
  Ground Truth & Google Cloud's Speech-to-Text & True & Predicted \\
   & Transcription & WER & WER\\
  \hline
\hline 
   one historian says that an event was produced by napoleon's power another that it was produced by alexander's & when is dorian says that an event was produced by napoleon's power another that it was produced by alexander's & 16.7 & 16.5 \\
\hline
   rynch watched dispassionately before he caught the needler jerking it away from the prisoner the man eyed him steadily and his expression did not alter even when rynch swung the off world weapon to center its sights on the late owner & wrench watch dispassionately before he caught a kneeler jerking it away from the prisoner the man i can steadily and his expression did not alter even when wrench swampy off world weapon to center its sights on the late owner & 21.9 & 22.1 \\
\hline
   of acting a father's part to augustine until he was fairly launched in life he had a child of his own & acting a father's part 2 augustine until he was fairly launched in life & 42.8 & 42.7 \\
\hline
   supported by an honorable name how could she extricate herself from this labyrinth to whom would she apply to help her out of this painful situation debray to whom she had run with the first instinct of a woman towards the man she loves and who yet betrays her & supported by an honorable name how could you extricate herself in this labyrinth to whom would she apply to help her out of this painful situation dubray to whom should run the first instinct of a woman towards the man she loves and who yep betrays her
 & 14.3 & 14.4 \\
\hline
    seventeen twenty four & 1724 & 100.0 & 30.7 \\
   \hline
 saint james's seven & st james 7& 100.0 & 32.1 \\
\hline
   mamma says i am never within & mama says i am never with him & 50.0 & 13.44 \\
   \hline
   \hline
\end{tabularx}
 \caption{Some examples of the proposed approach's WER Prediction}
 \label{tab:wer_examples}
\end{table*}

\section{WER-BERT}
\label{sec:wer_bert}
In this section we explain our proposed architecture WER-BERT, which is primarily made of four sub-networks. Our architecture is shown with details in Figure \ref{fig:arch}.

\textbf{Signal Sub-network}: \newcite{elloumi2018asr} use the raw signal of the audio clip to generate features such as MFCC and Mel Spectrogram. They are features commonly used in the design of ASR systems, particularly systems which use an acoustic model and furthermore these features aid their model performance. These signal features are passed through the m18 architecture \cite{dai2017very}. m18 is a popular deep convolutional neural network (CNN) used for classification with audio signal features. This CNN model has 17 convolutional+max pooling layers which is followed by global average pooling. L2 Regularization of $1e-4$ and Batch Normalization are added after each of the convolutional layers. %

\textbf{Numerical Features Sub-network: }\newcite{ali2018word} black box models had two major components --- text input and numerical features. These numerical features are important to the model as they contain information regarding the number of errors. For instance, in ASR systems, there are errors if a user speaks too fast or too slow and this is directly reflected in the duration and word count features. The numerical features we have used are Word Count, Grapheme Count and Duration. These features are concatenated and passed through a simple feed forward network which is used to upscale the numerical features fed into the model (from 3 to 32).

\label{sect:pdf}

\textbf{BERT: }Bi-directional Encoder Representations (BERT) \cite{devlin2018bert} is a pre-trained unsupervised natural language processing model. It is a masked language model which has been trained on a large corpus including the entire Wikipedia corpus. The transcription samples from the ASR system, are passed through the associated tokenizer which gives a contextual representation for each word. It also adds 2 special tokens --- the [CLS] token at the beginning and the [SEP] token at the end of the sentence. We have used the BERT-Large Uncased variant. The large variant has 24 stacked transformer \newcite{vaswani2017attention} encoders. It gives an output of the shape (Sequence Length X 1024) of which only the 1024 shaped output corresponding to the [CLS] token is used. In WER-BERT, BERT weights are fine tuned with rest of architecture during training.

\textbf{Feed Forward Sub-Network: }This sub-network is a deep fully connected network which is used to concatenate and process the features generated by the sub-networks predating it (BERT, numerical sub-network and the signal sub-network). It has 4 hidden layers (512, 256, 128 and 64 neurons) followed by the output softmax layer. Dropout regularization is added to prevent overfitting considering the large amount of parameters. To account for outputs from the eclectic sub-networks with disparate distributions, we further add Layer Normalization \cite{ba2016layer} before concatenation. Normalization is important to lessen the impact of any bias the network may learn towards one or the other representations.


\begin{figure*}
\centering
 \includegraphics[width=11cm]{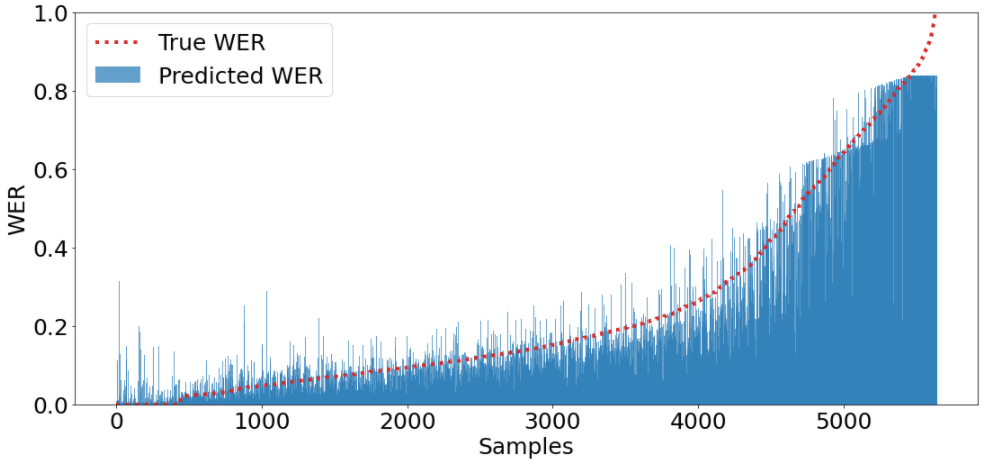}
 \caption{WER Predicted by WER-BERT SOTA model as compared to the True WER of all samples (Test 100hr)}
\label{fig:wer_pred}
\end{figure*}
\textbf{Distance Loss for Ordinal Classification: }Typical classification problems deal with classes which are mutually exclusive and independent such as sentiment prediction or whether an image is a cat or a dog. In such a setting, classification accuracy is the most important metric and there is no relation or relative ordering between the classes. However, e-WER in a classification setting is an ordinal classification problem \cite{frank2001simple}. Previous approaches which propose WER estimation as classification tasks ignore this idea \cite{elloumi2018asr,ali2018word,elloumi2019investigating}. While the classification accuracy is important, it is more important that given a sample is misclassified, the predicted label is close to the true label. For instance, if the true label corresponds to the WER class of 0.1, a prediction of 0.2 and a prediction of 0.7 are treated the same in the current classification scenario. Since we want the prediction to be as close as possible, if not exactly the same, we introduce a ``distance" loss which is $L_{custom}(s) = L_{\theta}(s) + \alpha * \gamma$ and $\gamma$ is as follows
\begin{dmath}
    \abs{y_{pred}(s) \cdot WER_{fixed} - y_{true}(s) \cdot WER_{fixed}}
\end{dmath}
where $s$ is a sample, $\alpha$ is a hyperparameter\footnote{Refer to Section A.1 of the appendix for tuning of the Distance Loss hyperparameter $\alpha$ } (we have used $\alpha = 50$ in our experiments), $L_{\theta}(s)$ is the cross entropy loss, $y_{true}(s)$ is a one hot vector representing the true WER class of $s$ and $y_{pred}$ is the estimated probability distribution of $s$ over all the classes output by the softmax of the classification model. 


\section{Experiments And Baselines}
\label{sec:baselines}
For each of the experiments below, the training is repeated for 10 runs and we report the average performance of all the runs on the test set. For all the experiments, we use Crossentropy as the loss function and $MAE$ of $WER$ as the evaluation metric.

\subsection{BOW}
\textbf{Bag of Words + Num. Feat.(Black Box): } Following \cite{ali2018word}, we build a black box model for the double task estimation of ERR and word count.
The word count estimation task was treated as a 46 class classification model with class 1 corresponding to word count of 2, class 46 with word count of 47 ($90^{th}$ percentile). Similarly, the ERR estimation task was modelled as a 20 class classification problem with class 0 corresponding to no errors and class 19 corresponding to 19 errors($90^{th}$ percentile). 
Both of these tasks are handled by logistic regression models which use Bag of Words features for both the words present in the sentence as well as the graphemes (monograms and bigrams) of the sentence. These features are concatenated with numerical features such as word count, grapheme count and the duration and then fed to a feedforward network. Dropouts are also added after each layer to prevent overfitting.

\begin{table*}[h]
\small
\centering
\begin{center}
\begin{tabular}{ll|l|l|l|l}
\hline \hline & \multirow{3}{*}{Model} & MAE & RMSE & MAE & RMSE\\ 
& & WER & WER & WER & WER\\ 
& & 100 hour & 100 hour & 10 hour& 10 hour\\ 
\hline
BoW &Bag of Words & \multirow{2}{*}{16.45} & \multirow{2}{*}{26.76} & \multirow{2}{*}{12.65} & \multirow{2}{*}{19.66}\\
& + Num. Feat.(Black Box) \cite{ali2018word} & & & \\
\hline
\multirow{5}{*}{CNN} & CNN-text (Balanced) & 14.44 & 18.9 & 13.08 & 18.79\\
&CNN-text + RAWSIG (Double Task) & 14.09 & 18.11 & 19.36 & 24.79\\
&CNN-text & \multirow{2}{*}{11.34} & \multirow{2}{*}{16.74} & \multirow{2}{*}{13.3} & \multirow{2}{*}{17.48}\\
& + RAWSIG \cite{elloumi2018asr}/(Imbalanced) & & &\\
&CNN-text + RAWSIG (Balanced) & 9.65 & 13.64 & 11.3 & 15.89\\ 
&CNN-text & \multirow{2}{*}{9.35} & \multirow{2}{*}{11.84} & \multirow{2}{*}{10.11} & \multirow{2}{*}{15.17}\\
& + MFCC + MELSPEC + RAWSIG / Signal Feat. (Balanced) & & & \\
\hline
 \multirow{5}{*}{WER-BERT} &BERT-\textit{large} & 12.04 & 17.98 & 11.19 & 16.38\\
&BERT-\textit{large} + Num. Feat.& 11.03 & 16.09 & 9.92 & 14.58\\
&BERT-\textit{large} + MFCC + MELSPEC + RAWSIG & 8.15 & 11.43 & 9.12 & 14.17\\ 
&BERT-\textit{large} + Num. Feat. + Signal Feat. & 6.91 & 10.01 & 8.89 & 13.52\\ 
&BERT-\textit{large} + Num. Feat. + Signal Feat. + Distance Loss & 5.98 & 8.82 & 7.37 & 12.67 \\
\hline
\hline
\end{tabular}
\end{center}
\caption{\label{tab:results} Performance Comparison. }
\end{table*}

\subsection{CNN}
Following, \newcite{elloumi2018asr} for e-WER. These CNN models use ASR transcript and signal features (Raw signal, MFCC and Mel Spectrogram) as inputs and CNN for learning features from the textual transcription itself.

\textbf{Text Input: }The text input is padded to T words (where T was taken as 50, the $95^{th}$ percentile of the sentence length) and was transformed into a matrix $Embeddings$ of the size $NXM$ where $M$ is the embedding size (300). These embeddings were obtained from GloVe \cite{pennington2014glove}. The CNN architecture is \newcite{kim2014convolutional}.

\textbf{Signal Inputs: }

\textit{RAWSIG: }This input is obtained by sampling the audio clip at 8KHz and max duration was set to 15s; shorter audio clips were padded while the longer ones were clipped. While using this with the M18 architecture, it was further down-sampled to 4KHz due to memory constraints and its dimension is $60000 X 1$.

\textit{MELSPEC: }This input was calculated with 96 dimensional vectors, each of which corresponds to a particular Mel frequency range. They were extracted every 10ms with an analysis window of 25ms and its dimension is $1501 X 96$.

\textit{MFCC:}
This input was calculated by computing 13 MFCCs every 10ms and its dimension is $1501 X 13$.
\newline
These signal features are used as an input to the M18 architecture \newcite{dai2017very} (refer to Section \ref{sec:wer_bert}: Signal sub-network). For joint use of both text and signal inputs, the outputs of the text and signal sub-networks are followed by a hidden layer (512 processing units) whose outputs are concatenated (with a dropout regularization of 0.1 being applied between the hidden layer and the concatenation layer). This is followed by 4 hidden layers (of 512, 256, 128 and 64 neurons) and the output layer with Dropout regularization added to prevent overfitting due to the large amount of parameters

\textbf{CNN-text (balanced): }While \newcite{elloumi2018asr} show that just CNN-text isn't enough to get good results. This experiment provides a baseline to compare against Bert-large since that does not have signal features as well. This model is trained in our balanced class setting with 15 classes explained in Section \ref{sec:balanced}.

\textbf{CNN-text + RAWSIG (Double Task): }
The best architecture in \newcite{elloumi2018asr} was trained in the double task setting proposed by \cite{ali2018word}. The ERR estimation was modelled as a 20 class task and the word count estimation was modelled as a 46 class task.

\textbf{CNN-text + RAWSIG \protect\cite{elloumi2018asr}: }
This was the best architecture proposed by \newcite{elloumi2018asr}. This is a single task approach which uses the imbalanced WER class distribution (classes corresponding to WER of 0, 0.25...1.5 WER).
\begin{figure*}
\centering
 \includegraphics[width=11cm]{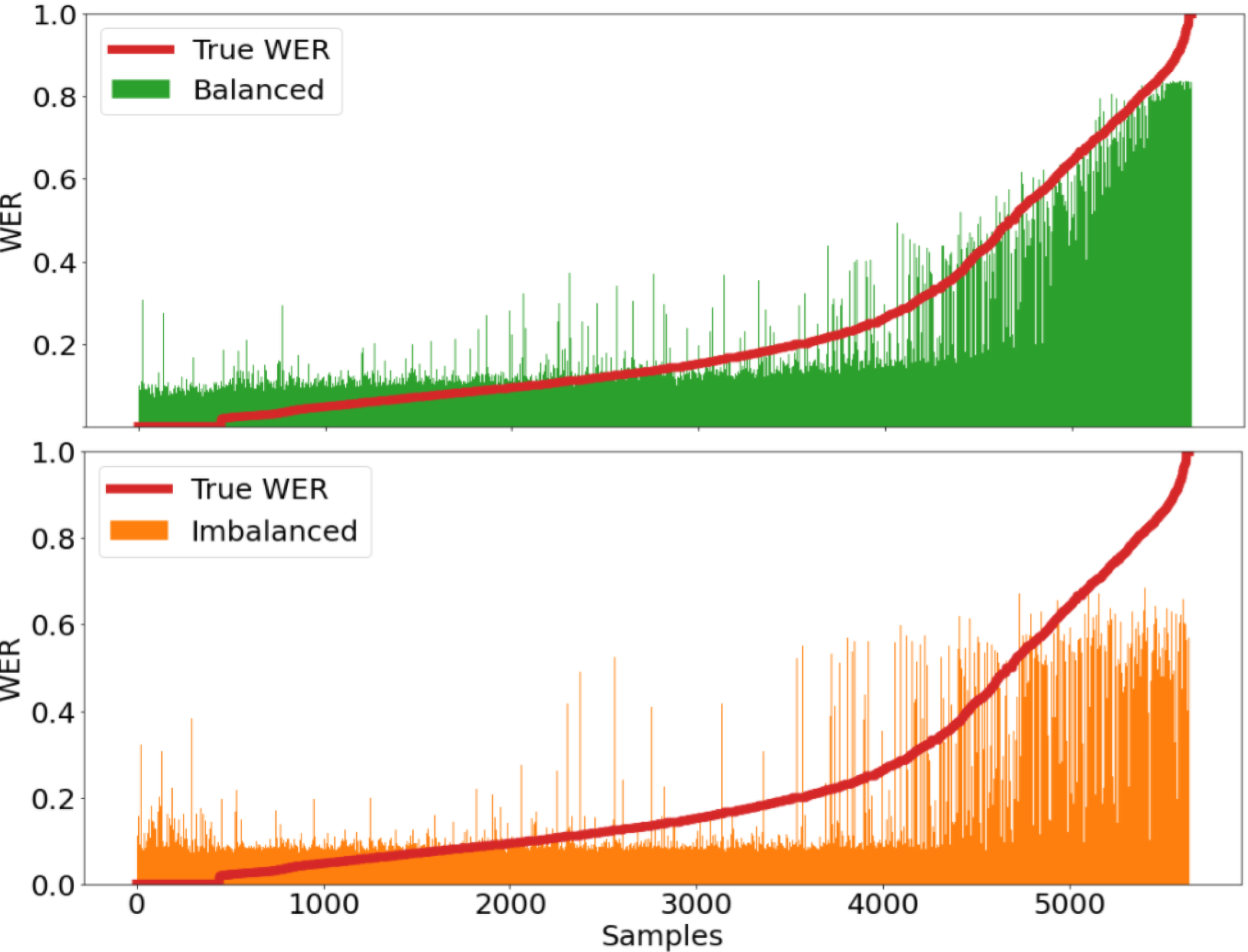}
 \caption{WER Predicted by the CNN + RAWSIG (Balanced) and CNN + RAWSIG \protect\cite{elloumi2018asr} (Imbalanced) as compared to True WER for all samples (Test 100hr)}
\label{fig:bal_imbal_wer}
\end{figure*}

\textbf{CNN-text + RAWSIG (Balanced): }
Same model as CNN-text + RAWSIG but trained in our balanced class setting with 15 classes explained in Section \ref{sec:balanced}.

\textbf{CNN-text + MFCC + MELSPEC + RAWSIG (Balanced): }
Instead of using only RAWSIG input, we pass all 3 signal features into their respective m18 models. The outputs of the KIM CNN model (for the text input) and each of these m18 models are concatenated and processed in same fashion as WER-BERT explained in Section \ref{sec:wer_bert}.
\begin{figure}
\centering
    \begin{subfigure}[b]{0.2\textwidth}
        \centering
        \includegraphics[width=3cm]{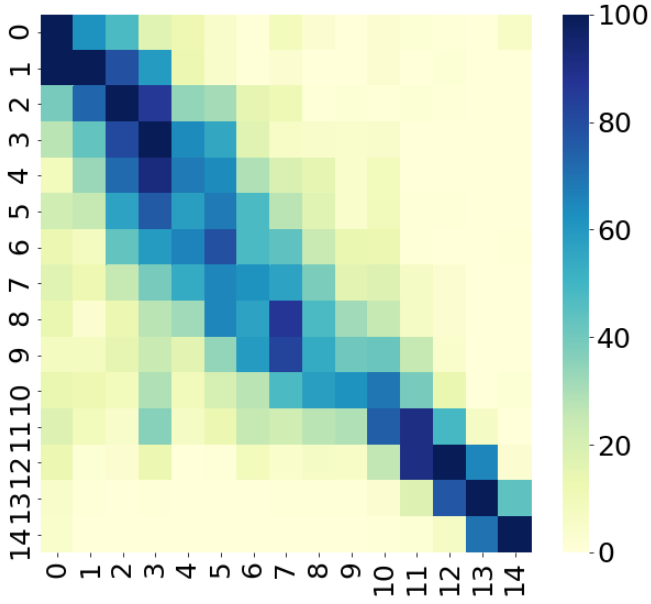}
        \caption{with Distance Loss}
        \label{fig:tiger}
    \end{subfigure}%
    \begin{subfigure}[b]{0.2\textwidth}
        \centering
        \includegraphics[width=3cm]{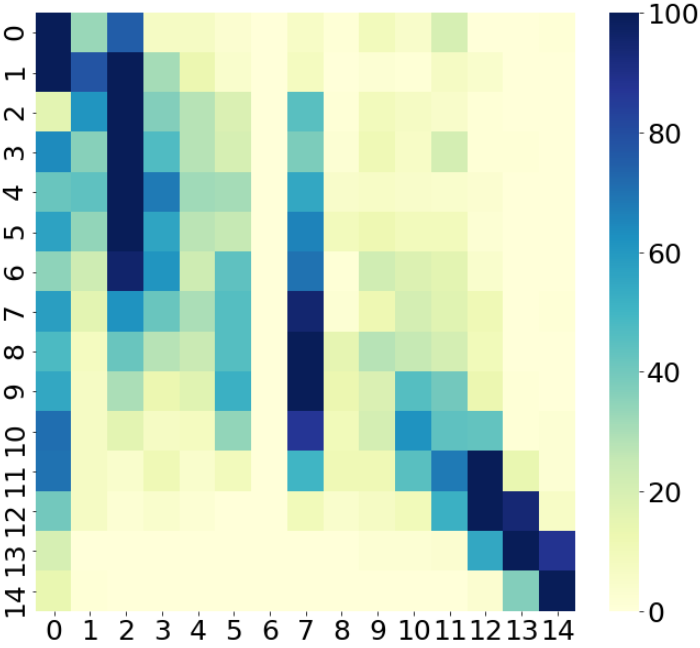}
        \caption{w/o Distance Loss}
        \label{fig:mouse}
    \end{subfigure}
    \caption{Confusion Matrices of the BERT-\textit{large} + All Features model\protect\footnote{For better visualization Confusion matrix values are scaled between 1 to 100}}\label{fig:confusion}
\end{figure}

\subsection{BERT architectures}
Experiments are carried with the architecture described in Section \ref{sec:wer_bert}. We carry out Ablation studies to identify important input features in isolation. We start with the full architecture shown in section \ref{sec:wer_bert} and subsequently remove sub-networks to compare performance. All the experiments except for the distance loss one are carried out with Crossentropy as the loss and $MAE$ of $WER$ as the evaluation metric and are trained with the balanced classes paradigm.

\section{Results and Discussion}
\label{sec:results}
Table \ref{tab:results} captures the results of proposed approach with previous approaches along with ablation studies for the proposed approach. While the WER of the 10 hour dataset is low ($\approx$10) and the 100 hour dataset is high ($\approx$20), we see that the proposed approach models both effectively. Figure \ref{fig:wer_pred} shows that WER-BERT's estimation of WER closely follows the true WER curve. Furthermore, due to the proposed balanced paradigm, it is able to predict well in the mid-high WER classes even with less samples in this region. Figure \ref{fig:bal_imbal_wer} shows the comparison of Balanced and Imbalanced class setting. Comparing Figure \ref{fig:wer_pred} and \ref{fig:bal_imbal_wer}, we see that the WER-BERT models much better in the lower and mid regions compared to the CNN balanced model.

\subsection{Effectiveness of WER-BERT and Ablation Studies}
The best WER-BERT model, with all Num., Signal feats. and distance loss outperforms other models in both MAE and RMSE metrics. In particular, we see that a BERT model outperforms the corresponding CNN model with the same inputs - 14.44 v/s 12.04 \& 13.08 v/s 11.19 for the CNN-text and BERT-large model in the 100 hour and 10 hour data respectively (and similar results for the CNN-text + signal feat and BERT-large + signal feat models). We credit effectiveness of BERT at its language model information and ability to identify improbable word sequences which often correlate with transcription errors. Models such as CNN, while being effective, lack the backing of a language model and hence fail to do the same. While this is important, it alone fails to beat the earlier approaches which utilise signal features i.e. Making a WER prediction from just the transcription is not enough. Signal features tells us how an utterance was spoken along with background noise. This contains valuable information such as signal noise which correlates with higher WER. The effectiveness of WER-BERT is particularly evident in Table \ref{tab:wer_examples} where we see that it is able to predict WER which is very close to the True WER. We hypothesize this is due to irregular word sequences such as ``..when is dorian says..'' or ``..the man i can steadily.'' being identified by BERT's language model. Furthermore, even when our model predicts a far off WER, it does so in a justified manner. The transcription ``1724" and ``st james 7'' give a high True WER score whereas WER-BERT identifies these sequences as correct and probable and gives low scores. Since WER matches strings exactly, it ends up giving a 100$\%$ error in this case due to library's text processing limitations. Note that while we use BERT, the language model itself can be any transformer, since they all have the common idea that LM backing helps identifying speech errors in ways CNN models can't.

\subsection{Effectiveness of proposed balancing setting}
We use the best architecture reported by \newcite{elloumi2018asr}: CNN-text + RAWSIG for comparing the three paradigms. Balanced outperforms the Double Task approach of \newcite{ali2018word} and Single Task approach \newcite{elloumi2018asr} (9.65 v/s 11.34 v/s 14.09 for the 100 Hour set). The class imbalance in their approach causes bias towards the largest classes during training, where most of the predictions end up. This especially harms the WER prediction in higher ranges where number of samples are less.
Figure \ref{fig:bal_imbal_wer} shows the performance difference between CNN + RAWSIG models in the new balanced paradigm and the imbalanced \newcite{elloumi2018asr} setting. Since the model encounters heavily imbalanced labels in the later setting, its predictions also reflect the same. There are only two kinds of predictions - one corresponding to the lowest WER class (also largest in number as seen in Figure \ref{fig:bal}) and another in the range of second largest class. The information of the other classes are mostly ignored due to being present in low numbers. Meanwhile, the balanced paradigm fits the slope of the curve after 40 WER well. The range of $>$ 40 WER is tough for the model to predict as the number of samples available in this region is lesser than samples in $<$ 40 WER (almost 70\% of data). Despite this, balanced paradigm effectively divides this area into adequate number of classes for good performance. While the performance in the mid region (20-40 WER) is poor for this model, custom loss and BERT model in WER-BERT take care of this as seen in Figure \ref{fig:wer_pred}.

\subsection{Effectiveness of Distance loss}
Addition of Distance loss to WER-BERT certainly improves the performance. In Figure \ref{fig:confusion} we see the confusion matrix of WER classification\footnote{Number of WER classes $K$ = 15} visualized with and without the custom loss. Due to the custom loss the models predicted class is much closer to the ground truth class represented by the diagonal of the matrix. This is due to distance loss' ability to penalize far off predictions which is lacking in typical classification loss functions such as cross-entropy. We see an improvement of nearly 1 MAE with this.

\section{Conclusion}
We propose WER-BERT for Automatic WER Estimation. While BERT is an effective model, addition of speech signal features boosts the performance. Phrasing WER classification as a ordinal classification problem by training using a custom distance loss encodes the information regarding relative ordering of the WER classes into the training. Finally, we propose a balanced paradigm for training WER estimation systems. Training in a balanced setting allows proposed model to predict WER adequately even in regions where samples are scarce. Furthermore, this balanced paradigm is independent of WER prediction model, ASR system or the speech dataset, making it efficient and scalable.

\bibliographystyle{acl_natbib}
\bibliography{eacl2021}

\appendix

\begin{table*}[]
\centering
\begin{tabular}{|l|l|l|l|l|}
\hline
Datasets    &  ERR &  Duration &  Length & True WER\\ \hline
Train 100hr & 8.16        & 12.74            & 34.84                   & 23.42     \\ \hline
Dev 100hr   & 8.13        & 12.70            & 34.64                   & 23.47     \\ \hline
Test 100hr  & 8.18        & 12.61            & 34.50                   & 23.70    \\ \hline
Train 10hr  & 2.65        & 7.29             & 20.44                   & 12.97      \\ \hline
Dev 10hr    & 2.28        & 7.15             & 19.96                   & 11.40      \\ \hline
Test 10hr   & 2.41        & 7.18             & 20.25                   & 11.91     \\ \hline
\end{tabular}
\caption{Averaged Statistics regarding the Train 100hr and Test 100hr dataset.}
\label{tab:data_details}
\end{table*}

\section{Hyperparameter Tuning Details}
In this section we explain the tuning and selection procedure of the proposed experiments' hyperparameters, namely the distance loss weight $\alpha$ and number of classes $K$. Furthermore, we also attempt to explain the behaviour of these variables on the performance in terms of MAE.
\subsection{Tuning Distance Loss Hyperparameter $\alpha$}
\begin{figure}
\centering
  \includegraphics[width=7.5cm]{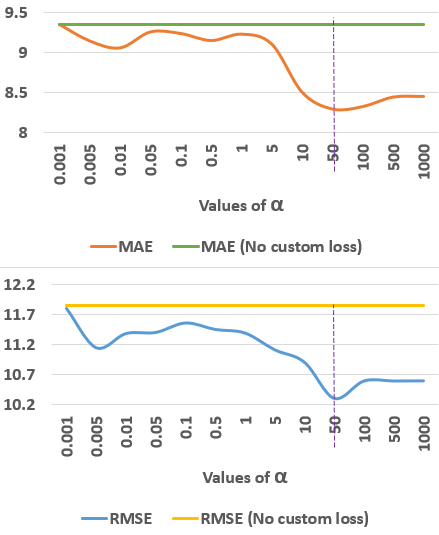}
  \caption{Distribution of MAE and RMSE for different values of $\alpha$ against no custom loss in WER-BERT. The values here are presented for the 100hr dataset.}
  \label{fig:alpha_tune}
\end{figure}

Figure \ref{fig:alpha_tune} shows the effect of varying $\alpha$ with effect on performance on the 100hr dataset with the WER-BERT model with and without custom loss. As evident,  we get the lower MAE and RMSE for $\alpha = 50$. Lower or higher $\alpha$ leads to decrement in performance. Recalling Equation 4 from the main paper:
\begin{equation}
    L_{custom}(s) =  L_{\theta}(s) + \alpha * \gamma
\end{equation}
where  $\gamma$ is:
\begin{dmath}
        \abs{y_{pred}(s) \cdot WER_{fixed} - y_{true}(s) \cdot WER_{fixed}}
\end{dmath}
For low values of $\alpha$ (0.0001 to 0.001), in the above equations  $L_{custom}(s)$ becomes equal to just  $L_{\theta}(s)$ which the classification cross entropy loss. Hence, essentially low $\alpha$ performance tends to the line of No custom loss WER-BERT performance. On the other hand, higher $\alpha$ values weight the custom loss too much more than the cross entropy loss. While the decrement in performance isn't very high, moderate adverse effects still show that, regular classification cross entropy loss is required for WER-BERT training.

\begin{figure}

\centering
  \includegraphics[width=7.5cm]{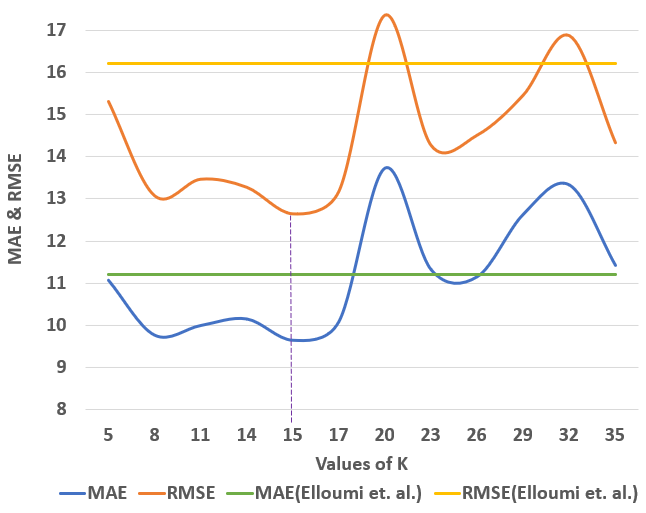}
  \caption{Distribution of MAE and RMSE for different values of $K$ against \protect\newcite{elloumi2018asr} approach. The values here are presented for the CNN + RAWSIG model on the 100hr dataset.}
  \label{fig:k_tune}
\end{figure}

\subsection{Tuning Class Hyperparameter $K$}

Figure \ref{fig:k_tune} shows the effect on performance of the CNN + RAWSIG model for varying MAE. We get the minimum MAE and RMSE at $K=15$.  Largely speaking, for lower $K$, RMSE and MAE suffer because the new classes no more accurately depicts the true WER. For example, in Figure 1 of main paper, we can see that if $K=3$ or $4$, WERs in a wide range from 40 to 100 will be clubbed into a single class despite their huge differences. This will further make it hard for the model to predict WERs in this range. 

On the other hand, higher $K$ yields new classes accurate to true WER especially in the higher ranges(50 to 100), but at the cost of class size. For example at K=30, each class will have about 900 (28000/30)  samples.Furthermore, since True WER is a highly imbalanced variable, there exists a long tail in the lower WER regions in Figure 1 of main paper. This WER region is divided into many classes, despite the WER range being largely small. For example for $K>30$ the model will be forced to distinguish between nearly four classes with arbitrary samples of all 0 True WER. Just for $K=15$, model encounters two classes which are actually both 0 WER. This number of redundant classes with arbitrary samples will only increase as $K$ increases.

Furthermore, for a range of values of $K$, we see that the balanced CNN + RAWSIG model still outperforms the CNN + RAWSIG \cite{elloumi2018asr}'s performance. This further reinforces the efficacy of the balanced paradigm even with its senstivity to a hyperparameter.

\section{Dataset}

Additional dataset details are present in Table \ref{tab:data_details}.

\end{document}